\documentclass{llncs}
\usepackage{makeidx}

\usepackage[misc]{ifsym}

\usepackage{times}

\usepackage{graphicx} 
\usepackage{subfigure} 
\usepackage{color}
\usepackage{mathtools}
\usepackage{algorithm}
\usepackage{algorithmic}
\usepackage{amsmath}
\usepackage{amssymb}
\usepackage{tikz}
\usetikzlibrary{bayesnetLJ}

\DeclareMathOperator*{\argmax}{argmax} 

\def\bw{{\bf w}}
\def\bW{{\bf W}}

\def\bX{{\bf X}}

\def\bS{{\bf S}}
\def\bw{{\bf w}}
\def\bg{{\bf g}}
\def\ba{{\bf a}}
\def\bd{{\bf d}}

\newcommand\eg{{\em e.g., }}
\newcommand\ie{{\em i.e., }}

\def\btheta{{\boldsymbol\theta}}

\def\bpi{{\boldsymbol\pi}}

\def\balpha{{\boldsymbol\alpha}}
\def\bbeta{{\boldsymbol\beta}}
\def\bgamma{{\boldsymbol\gamma}}
\def\bdelta{{\boldsymbol\delta}}

\titlerunning{Identifying Readers and Assessing Text Comprehension from Eye Movements}

\begin{document} 

\title{A Discriminative Model for Identifying Readers and Assessing Text Comprehension from Eye Movements}

\author{Silvia Makowski\inst{1}\thanks{Joint first authorship}\and
	Lena A. J\"ager\inst{1,2,3}$^\star$\and
	Ahmed Abdelwahab\inst{1,4}\and
	\\Niels Landwehr\inst{1,4} \and
	Tobias Scheffer\inst{1}}

\authorrunning{Makowski et al.} 

\institute{University of Potsdam, Department of Computer Science, 
Potsdam, Germany\\
\and University of Potsdam, Department of Linguistics, 
Potsdam, Germany 
\and Weizenbaum Institute for the Networked Society, Berlin, Germany\\
\and Leibniz Institute for Agricultural Engineering and Bioeconomy, 
Potsdam, Germany\\
	\email{\{smakowsk, lejaeger, tobias.scheffer\}@uni-potsdam.de,  \{NLandwehr, AAbdelwahab\}@atb-potsdam.de}
	}

\maketitle              

\begin{abstract}
We study the problem of inferring readers' identities and estimating their level of text comprehension from observations of their eye movements during reading. We develop a generative model of individual gaze patterns (\textit{scanpaths}) that makes use of lexical features of the fixated words. Using this generative model, we derive a Fisher-score representation of eye-movement sequences. 
We study whether a Fisher-SVM with this Fisher kernel and several reference methods are able to identify readers and estimate their level of text comprehension based on eye-tracking data. While none of the methods are able to estimate text comprehension accurately, we find that the SVM with Fisher kernel excels at identifying readers.
	
	\keywords{Fisher kernel, eye movements, reader identification}
\end{abstract}

\section{Introduction}
During reading, the eye proceeds in a series of rapid movements, called saccades, instead of smoothly wandering over the text. Between two saccades, the eye remains almost still for about 200 to 300 milliseconds on average, fixating a certain position in text to obtain visual input. Saccades serve as a relocation mechanism of the eye moving the focus on average seven to nine characters wide from one fixation position to the next. Eye movements during reading are driven by complex cognitive processes involving vision, attention, language and oculomotor control \cite{kliegl2006,Rayner1998}. Since a reader's eye movement behavior is precisely observable and reflects the interplay of internal processes and external stimuli for the generation of complex action \cite{kliegl2006}, it is a popular research subject in cognitive psychology. 

One common insight of various studies in the field is that eye movement patterns vary significantly between individuals \cite{erdmann1898,huey1908,Afflerbach2015}. This property makes them interesting for biometrics. Indeed, identification based on eye movements during reading may offer several advantages in many application areas. Users can be identified unobtrusively while having access to a document they would read anyway, which saves time and attention. 
For biometric identification during reading, nearest-neighbor~\cite{Holland2011} and generative probabilistic models~\cite{Landwehr2014,AbdelwahabEtAl2016} of eye-gaze patterns have been explored. 

Eye movements are believed to mirror different levels of comprehension processes involved in reading~\cite{JustCarpenter1980}. Experimental studies have shown that reader's fixations are influenced by syntactic comprehension~\cite{Frazier1982}, semantic plausibility~\cite{Staub2007}, background knowledge~\cite{Kaakinen2007},  text difficulty, and inconsistencies~\cite{Rayner2006}. 
These findings motivate our goal of estimating readers' levels of text comprehension based on their eye-gaze. 

Gaze patterns, also referred to as \textit{scanpaths}, that occur during reading are sequences of fixations and saccades. One can easily extract vectors of aggregated distributional features that standard learning algorithms can process~\cite{Holland2011}---for instance, the average fixation duration and saccade amplitude---albeit at a great loss of information. Generative graphical models~\cite{Landwehr2014} allow to infer the likelihood of a scanpath under reader-specific model parameters. However, since both identification and assessing text comprehension are discriminative tasks, it appears plausible that discriminatively trained models would be better suited to this task. Classifying sequences by a discriminative model involves engineering a suitable sequence kernel or other form of data representation. Recurrent neural networks tend to work well for problems for which large data collections are available to train  high-capacity models. By contrast, eye movement data cannot be collected at a large scale, because their collection requires laboratory equipment and test subjects. We therefore focus on the development of a suitable sequence kernel. We will follow the approach of Fisher kernels because it allows us to use background knowledge in the form of a plausible generative model as the representation of scanpaths. 

Based on an existing generative model by \cite{Landwehr2014}, we develop a model that takes into account lexical features of the fixated words to generate a scanpath. This model is then used to map eye scanpaths into Fisher score vectors. We classify with an SVM and the Fisher kernel function such that we exploit both the advantages of generative modeling and the strengths of discriminative classification. 

The rest of this paper is organized as follows. Section~\ref{sec:related} reviews related work. Section 3 introduces the problem setting and notation. In Section~\ref{sec:method}, we develop a generative model of scanpaths that takes into account the lexical features of the fixated word, and derive the corresponding Fisher kernel in Section~\ref{Sec:Fisher_kernels}. In Section 6, we show the empirical evaluation; Section 7 concludes.

\section{Related Work}\label{sec:related}

Eye movements are assumed to mirror cognitive processes involved in reading~\cite{JustCarpenter1980}. A large body of psycholinguistic evidence shows that language comprehension processes at the syntactic, semantic and pragmatic level are significant predictors for a reader's fixation durations and saccadic behavior~\cite{Frazier1982,Rayner1994,Rayner1998,Rayner2006,Clifton2007}. 

For our purposes, effects 
of higher level text comprehension (\ie on the level of the discourse) on a reader's eye movements  are most relevant. For example, it has been shown that conceptual difficulty of a text leads to a larger proportion of regressions, an increase in fixation durations, and a decrease in saccade amplitudes~\cite{Jacobson1979,RaynerPollatsek1989}. Rayner et al.~\cite{Rayner2006} show that higher global and local discourse difficulty of a text increases the number and average duration of fixations as well as the proportion of regressive saccades. Semantically impossible or implausible words have been shown to increase the first-pass reading time and the total reading time of a word, respectively \cite{Rayner2004,Warren2008}. 
Moreover, background knowledge decreases both the sum of all fixation durations on a word when reading it for the first time and the proportion of skipped words~\cite{Kaakinen2007}.

Existing attempts to exploit this eye-mind connection and actually use a reader's eye movements to predict text comprehension have crucial limitations. Copeland et al.~\cite{Copeland2013,Copeland2014b,Copeland2014a} use the saccades between a comprehension question and the text as a feature to predict the response accuracy on this very question. Hence, these models are not trained to infer reading comprehension from the eye movements while reading a text, as claimed by the authors, but rather predict response accuracy on a question from the answer-seeking eye movements of the user. Indeed, the practical relevance of predicting text comprehension from reading is that no questions would be needed anymore to assess a reader's comprehension of a text.
Underwood et al.~\cite{Underwood1990} also claim to predict text comprehension from a reader's fixation durations. However, they use the same data for training and testing their model.


Compared to the usually rather small size of the effects reflecting cognitive processes, individual variability of eye movements in reading is very large. This has been observed 
consistently in the psychological literature~\cite{huey1908,dixon1951studies,rayner2012psychology}. The idea behind eye movements as a biometric feature is to exploit this individual variability. 
Some biometric studies are based on eye movements observed in response to an artificial visual stimulus, such as a 
moving~\cite{komogortsev2010biometric,rigas2012human,zhang2012on} or fixed~\cite{bednarik2005eye} dot on a computer screen, or a specific image stimulus~\cite{rigas2012biometric}.
Other studies, like our paper, focus on the problem of identifying subjects while they process an arbitrary stimulus, which has the advantage that the identity 
can be inferred unobtrusively during routine access to a device or document. Holland and Komogortsev study identification of subjects from eye movements on arbitrary text, based on aggregated statistical features such as the average fixation duration and average saccade amplitude~\cite{Holland2011}. 
Rigas et al. extend this approach with additional dynamic saccadic features~\cite{rigas2016biometric}. 
However, by reducing observations to a small set of real-valued features, much of the information in eye movements is lost.
Landwehr et al.~\cite{Landwehr2014} show that by fitting subject-specific generative probabilistic models to eye movements, 
much higher identification accuracies can be achieved.
They develop a parametric generative model
~\cite{Landwehr2014}; as this model serves as a starting point for our method, details are given in Section~\ref{sec:landwehr_model}.
Abdelwahab et al.~\cite{AbdelwahabEtAl2016} extend this model to a fully Bayesian approach, in which distributions are defined by nonparametric densities inferred under a Gaussian process prior that is centered at the gamma family of distributions~\cite{AbdelwahabEtAl2016}. Both methods serve as reference methods in our experiments.

\section{Problem Setting}
\label{sec:problem_setting}
\sloppy

When reading a text $\bX$, a reader generates a scanpath that is given by a sequence
\mbox{$\bS = ((q_1, d_1),\ldots,(q_T,d_T))$} of fixation positions $q_t$ (position in text that was fixated, measured in characters) 
and fixation durations $d_t$ (measured in milliseconds). 
This scanpath can be observed with an eye-tracking system. 

Each word fixated at time $t$ possesses lexical features that can be aggregated into a vector $\bw_t$. Some of the models that we will study will allow the distributions of saccade amplitudes and durations to depend on such lexical features. Lexical features---for instance, word frequency or part of speech---are derived from the text $\bX$ itself. 

We study the problems of reader identification and assessing text comprehension. In reader identification, the model output $y$ is the conjectured identity of the reader that generates scanpath $\bS$ for text $\bX$, from a set of individuals that are known at training time. In assessing text comprehension, the model output is the conjectured level of the reader's comprehension $y$ of text $\bX$. In order to annotate training and evaluation data, the ground-truth level of text comprehension can be determined, for instance, by a question-answering protocol carried out after reading. In an actual application setting, no comprehension questions are asked. 

In both settings, training data consists of a set $\mathcal{D} = \{(\bS_1,\bX_1,y_1),...,(\bS_n,\bX_n,y_n)\}$
of scanpaths $\bS_1,...,\bS_n$ that have been obtained from subjects reading texts $\bX_1,...,\bX_n$, annotated with labels $y_1,...,y_n$. 
%

\fussy

\section{Generative Models of Scan Paths}
\label{sec:method}
Landwehr et al.~\cite{Landwehr2014} 
define a parametric model $p(\mathbf{S}|\mathbf{X},\btheta)$ of scanpaths given a text $\bX$. 
Fitting this model to the subset of scanpaths and texts ${\mathcal{\bar D}}_y = \{(\bS_i,\bX_i)|(\bS_i,\bX_i,y_i)\in {\mathcal{D}}, y_i=y\}$ in the training data generated by reader $y$ yields reader-specific 
models $p(\mathbf{S}|\mathbf{X},\btheta_y)$.
At application time, the prediction for a scanpath $\bS$ on a novel text $\bX$ can be obtained as $y^* = \argmax_y p(\mathbf{S}|\mathbf{X},\btheta_y)$.
We first review this generative model~\cite{Landwehr2014}, and then develop it into a generative model of scanpaths that takes into account the lexical features of the fixated words in Section~\ref{Sec:Generative_Model_Lexical_Features}. In Section~\ref{Sec:Fisher_kernels}, we derive the Fisher kernel and arrive at a discriminative model.

\sloppy
\subsection{The Model of Landwehr et al., 2014}
\label{sec:landwehr_model}
This section presents a slightly simplified version of the generative model of scanpaths $p(\mathbf{S}|\mathbf{X},\btheta)$~\cite{Landwehr2014}.
It reflects how readers generate fixations while reading a text and models the type and amplitude of saccadic movements and fixation durations. 
The joint distribution over all fixation positions and durations is assumed to factorize as
\begin{align}
\label{eq: Landwehr Joint Distribution}
p(q_1,\dots,q_T,d_1,\dots,d_T|\bX,\btheta) =p(q_1,d_1|\bX,\btheta)\prod_{t=1}^{T-1} p(q_{t+1},d_{t+1}|q_{t},\bX,\btheta).
\end{align}
To model the conditional distribution $p(q_{t+1},d_{t+1}|q_{t},\bX,\btheta)$ of the next fixation position and duration given the current fixation position, 
the model distinguishes five \emph{saccade types} $u$: a reader can 
refixate the current word at a character position before the current position ($u=1$), refixate the current word at a position after the current position ($u=2$), fixate the next word in the text ($u=3$), move the fixation to a word after the next word ($u=4$), or regress to fixate a word occurring earlier in the text ($u=5$).
At each time $t$, the model first draws a saccade type
\begin{equation}
\label{eq: Landwehr Saccade Type}
u_{t+1} \sim p(u|\bpi)= \mathrm{Mult}(u|\bpi)\\
\end{equation}
from a multinomial distribution. It then draws a (signed) saccade amplitude\footnote{Throughout our work, saccade amplitude is measured in number of characters as this metric is relatively insensitive to differences in the eye-to-screen distance, which might become relevant for practical applications of the model \cite{MorrisonRayner1981}.} \mbox{$a_{t+1} \sim p(a| u_{t+1},\balpha,\bbeta)$} from type-specific gamma distributions; that is, 
\begin{align}
&p(a|u_{t+1}=u,\balpha,\bbeta) = \mathcal{G}(a|\alpha_u,\beta_u) \text{ for $u \in \{2,3,4\}$} \label{eq:amplitude_positive}\\
&p(a|u_{t+1}=u,\balpha,\bbeta) = \mathcal{G}(-a|\alpha_u,\beta_u) \text{ for $u \in \{1,5\}$} \label{eq:amplitude_negative}
\end{align}
where $\balpha=\{\alpha_u|u \in \{1,...,5\}\}$, $\bbeta=\{\beta_u|u \in \{1,...,5\}\}$ and $\mathcal{G}(\cdot|\alpha,\beta)$ is the gamma distribution parameterized by
shape $\alpha$ and scale $\beta$. The current fixation position is then updated as $q_{t+1} = q_t + a_{t+1}$.
The model finally draws the fixation duration $d_{t+1} \sim p(d| u_{t+1},\bgamma,\bdelta)$, also from type-specific gamma distributions
\begin{align}
&p(d|u_{t+1}=u,\bgamma,\bdelta) = \mathcal{G}(d|\gamma_u,\delta_u) \text{ for $u \in \{1,2,3,4,5\}$} \label{eq:duration}
\end{align}
where $\bgamma=\{\gamma_u|u \in \{1,...,5\}\}$, $\bdelta=\{\delta_u|u \in \{1,...,5\}\}$.
All parameters of the model are aggregated into a parameter vector $\btheta$.

The difference between this simplified variant and the original model~\cite{Landwehr2014} is that the original model truncates the gamma distributions in order to fit within the limits of the text interval defined by the saccade type; for instance, to the currently fixated word for refixations. Since this truncation causes unsteadiness of the Fisher scores, we instead let the amplitudes be governed by regular gamma distributions with scale parameter $\alpha_u$ or $\gamma_u$ and shape parameter $\beta_u$ or $\delta_u$. Furthermore, Landwehr et al. distinguish the same 
five saccade types for modeling saccade amplitude, but only four saccade types for modeling fixation durations, while we distinguish five saccade types for both distributions.

\fussy

\subsection{Generative Model with Lexical Features}
\label{Sec:Generative_Model_Lexical_Features}
We extend the model presented in Section~\ref{sec:landwehr_model} by allowing the distributions of fixation durations and saccade amplitudes to depend on lexical features $\bw_t$ of each fixated word. 

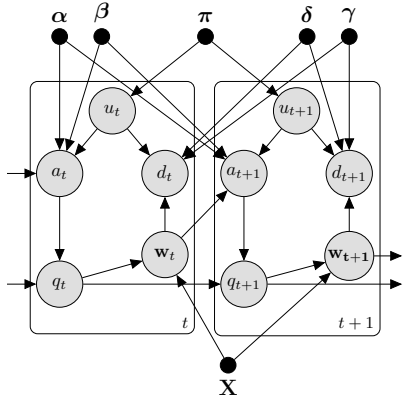
\begin{figure}
\centering
\scalebox{0.7}{
\begin{tikzpicture}
    \node[obs, minimum size=25pt] (u1) {$u_t$};%
    \node[obs, below=of u1, minimum size=25pt, xshift=-1cm,  yshift=20] (a1) {$a_t$};%
    \node[obs, below=of u1, minimum size=25pt, xshift=1cm, yshift=20] (d1) {$d_t$};
    \node[obs,below=of a1, minimum size=25pt, yshift=-6] (q1) {$q_t$};
    \node[obs ,below=of d1, minimum size=25pt, yshift=10] (w1) {$\bw_t$};

    \node[obs, xshift=3.5cm, minimum size=25pt] (u2) {$u_{t+1}$};%
    \node[obs, below=of u2, minimum size=25pt, xshift=-1cm, yshift=20] (a2) {$a_{t+1}$};
     \node[obs, below=of u2, minimum size=25pt, xshift=1cm, yshift=20] (d2) {$d_{t+1}$};
    \node[obs,below=of a2, minimum size=25pt, yshift=-6] (q2) {$q_{t+1}$};
    \node[obs ,below=of d2, minimum size=25pt, yshift=10] (w2) {$\bf w_{t+1}$};

    \node [const, above=of d1,  yshift=1cm, xshift=-1.2cm, label=above:\large$\boldsymbol\beta$] (beta) {};
    \node [const, above=of d1,  yshift=1cm, xshift=-2cm, label=above:\large$\boldsymbol\alpha$] (alpha) {};
    \node [const, above=of a2,  yshift=1cm, xshift=1.2cm, label=above:\large$\boldsymbol\delta$] (delta) {};
    \node [const, above=of a2,  yshift=1cm, xshift=2cm, label=above:\large$\boldsymbol\gamma$] (gamma) {};
    \node [const, above=of a1,  yshift=1cm, xshift=2.77cm, label=above:\large$\boldsymbol\pi$] (pi) {};

    \node[const, below=of w1,  yshift=-0.5cm, xshift=1.2cm, label=below:\large$\mathbf X$] (X) {};

    \plate  {plate1}  {(u1) (q1) (a1) (d1) (w1)} {$t$}; %
    \plate {plate2} {(u2) (q2) (a2) (d2) (w2)} {$t+1$}; %

    \edge {u1} {d1,a1} 
    \edge {a1} {q1} 
    \edge {q1} {w1} 
    \edge {w1} {d1}

    \edge {w1}{a2}
    \edge {q1}{q2}

    \edge {u2} {d2,a2}
    \edge {a2} {q2} 
    \edge {q2} {w2} 
    \edge {w2} {d2}

    \edge{alpha}{a1,a2}
    \edge{beta}{a1,a2}
    \edge{gamma}{d1,d2}
    \edge{delta}{d1,d2}
    \edge{pi}{u1,u2}
    \edge{X}{w1,w2}

    \draw[<-] (q1) --++(-1cm,0);
    \draw[->] (q2) --++(0:3cm);
    \draw[<-] (a1) --++(-1cm,0);
    \draw[->] (w2) --++(0:1cm);
     \end{tikzpicture}
     }
 \caption{Plate notation of the generative model for scanpaths with lexical features.}
 \label{fig:graphical_model}
\end{figure}

Let the random variable $\bw_t$ denote a vector of features of the word that is fixated at time step $t$, 
such as word frequency or length (Section~\ref{sec:empirical_study} gives more details on the features under study).
We allow these features to influence the scale and shape of the gamma distributions from which the saccade amplitudes and fixation durations are generated. Hence, we model the scale and shape parameters $\alpha_u, \beta_u, \gamma_u, \delta_u$ in Equations~\ref{eq:amplitude_positive},~\ref{eq:amplitude_negative} and~\ref{eq:duration} as linear regressions on the word features ${\mathbf w}_t$ with an exponential link to ensure positivity of the gamma parameters. That is, we replace Equations~\ref{eq:amplitude_positive},~\ref{eq:amplitude_negative} and~\ref{eq:duration} by
\begin{align}
&p(a|u_{t+1}=u,\bw_t,\balpha,\bbeta) = \mathcal{G}(a|\exp(\balpha_u^T \bw_t), \exp(\bbeta_u^T \bw_t)) \text{ for $u \in \{2,3,4\}$} \label{eq:amplitude_positive_feat},
\\
&p(a|u_{t+1}=u,\bw_t\balpha,\bbeta) = \mathcal{G}(-a|\exp(\balpha_u^T \bw_t), \exp(\bbeta_u^T \bw_t)) \text{ for $u \in \{1,5\}$} \label{eq:amplitude_negative_feat}
\end{align}
and
\begin{align}
&p(d|u_{t+1}=u,\bw_t,\bgamma,\bdelta) = \mathcal{G}(d|\exp(\bgamma_u^T \bw_{t+1}), \exp(\bdelta_u^T \bw_{t+1})) \text{ for $1\leq u \leq 5$} \label{eq:duration_feat}.
\end{align}

%

Note that $\balpha_u,\bbeta_u,\bgamma_u,\bdelta_u$ are now vectors of regression weights from which the respective gamma parameters are computed, 
which are aggregated into the parameterizations $\balpha=\{\balpha_u|u \in \{1,...,5\}\}$, $\bbeta=\{\bbeta_u|u \in \{1,...,5\}\}$, $\bgamma=\{\bgamma_u|u \in \{1,...,5\}\}$, and $\bdelta=\{\bdelta_u|u \in \{1,...,5\}\}$. Figure~\ref{fig:graphical_model} shows a graphical model representation.

\sloppy 
\subsection{Parameter Estimation}
\label{sec:parameter_estimation}
Given a set of scanpaths and texts ${\mathcal{\bar D}} = \{(\bS_i,\bX_i)\}$, 
model parameters can be estimated by maximum
likelihood. In a generative setting, models for a specific reader $y$ or a specific discrete competence level $y$ can be be estimated on a data subset ${\mathcal{\bar D}}_y = \{(\bS_i,\bX_i)|(\bS_i,\bX_i,y_i)\in {\mathcal{D}}, y_i=y\}$.
For the discriminative setting we develop in Section~\ref{Sec:Fisher_kernels}, generative parameters are estimated on all training data ${\mathcal{\bar D}} = \{(\bS_i,\bX_i)|(\bS_i,\bX_i,y_i)\in {\mathcal{D}}\}$, and a Fisher score representation is derived from this generative model.
We optimize a regularized maximum likelihood criterion
\begin{equation}
\label{eq:maximum_likelihood}
\btheta^* = \argmax_{\btheta} \sum_{i=1}^k \ln p(\bar\bS_i|\bar\bX_i,\btheta) - \lambda \Omega(\btheta).
\end{equation}
Given $\bar{\mathcal{D}}$, all fixation positions $q_t$, saccade types $u_t$ and word features $\bw_t$ are known. Equation~\ref{eq:maximum_likelihood}
thus factorizes into separate likelihood terms depending on saccade type, amplitude, and duration parameters:
\begin{multline}
\btheta^* = \argmax_{\bpi,\balpha,\bbeta,\bgamma,\bdelta} \bigg(\sum_{i=1}^k \sum_{t=1}^{T_i} \ln \mathrm{Mult}(u^{(i)}_t| \bpi) 
 \\+ \sum_{i=1}^k \sum_{t=1}^{T_i} \ln p(a^{(i)}_t | u^{(i)}_t,\bw^{(i)}_t,\balpha,\bbeta) - \lambda \Omega(\balpha,\bbeta)  \\
+ \sum_{i=1}^k \sum_{t=1}^{T_i} \ln p(d^{(i)}_t | u^{(i)}_t,\bw^{(i)}_t,\bgamma,\bdelta) - \lambda \Omega(\bgamma,\bdelta)\bigg) \label{eq:likelihood_factorized}
\end{multline}
where $u^{(i)}_t$, $a^{(i)}_t$, $d^{(i)}_t$, and $\bw^{(i)}_k$ denote saccade types, amplitudes, fixation durations, and word features in $(\bar{\bS}_i,\bar{\bX}_i)$, the number of fixations in sequence $\bar{\bS}_i$ is written as $T_i$, and we have split up the regularizer into separate regularizers $\Omega(\balpha,\bbeta)$ and $\Omega(\bgamma,\bdelta)$ (parameter $\bpi$ is not regularized).
Equation~\ref{eq:likelihood_factorized} can be optimized independently in saccade type parameters $\bpi$, amplitude parameters $\balpha,\bbeta$, and duration parameters $\bgamma,\bdelta$.
Optimization in $\bpi$ is straightforward. Because given $\bar{\mathcal{D}}$, saccades types are known, amplitude parameters 
can be optimized independently for each saccade type; that is, optimization is independent for each $\balpha_u,\bbeta_u$. Let $u \in \{1,...,5\}$, then
\begin{align}
(\balpha_u^*,\bbeta_u^*) &= \argmax_{\balpha_u,\bbeta_u} \sum_{i=1}^k \sum_{1 \leq t  \leq T_i:u^{(i)}_t=u}\!\!\!\!\! \ln p(a^{(i)}_t| u^{(i)}_t=u, \bw^{(i)}_t, \balpha_u, \bbeta_u) 
- \lambda \Omega(\balpha_u,\bbeta_u)\notag \\
& = \argmax_{\balpha_u,\bbeta_u} \sum_{i=1}^k  \sum_{1 \leq t \leq T_i:u^{(i)}_t=u}\!\!\!\!\!\ln \mathcal{G}(|a^{(i)}_t| | \exp(\balpha_u^\top \mathbf{w}^{(i)}_t), \exp(\bbeta_u^\top \mathbf{w}^{(i)}_t))\notag \\
\vspace{-1cm}& \hspace{40mm} - \lambda \sum\nolimits_{m=0}^{M-1}\exp(\alpha_{u,m})+\exp(\beta_{u,m}) \label{eq:factorization_over_u}
\end{align}
where $M$ is the number of lexical features used to predict the gamma parameters (including a bias),  and $\alpha_{u,m}$, $\beta_{u,m}$ denote the $m$-th element of parameter
vectors $\balpha_u$, $\bbeta_u$ respectively. Note that as the linear regression on the word features is scaled using an exponential function (Equations~\ref{eq:amplitude_positive_feat},~\ref{eq:amplitude_negative_feat}), we use an exponential regularizer $\Omega(\balpha,\bbeta)$.
Analogously, for fixation durations,
\begin{align}
(\bgamma_u^*,\bdelta_u^*) &= \argmax_{\bgamma_u,\bdelta_u} \sum_{i=1}^k \sum_{1 \leq t  \leq T_i:u^{(i)}_t=u}\!\!\!\!\! \ln p(d^{(i)}_t| u^{(i)}_t=u, \bw^{(i)}_t, \bgamma_u, \bdelta_u)
- \lambda \Omega(\bgamma_u,\bdelta_u)\notag\\
& = \argmax_{\bgamma_u,\bdelta_u} \sum_{i=1}^k  \sum_{1 \leq t \leq T_i:u^{(i)}_t=u}\!\!\!\!\!\ln \mathcal{G}(d^{(i)}_t | \exp(\bgamma_u^\top \mathbf{w}^{(i)}_t), \exp(\bdelta_u^\top \mathbf{w}^{(i)}_t)) \notag\\
& \hspace{40mm} - \lambda \sum\nolimits_{m=0}^{M-1}\exp(\gamma_{u,m})+\exp(\delta_{u,m}).
\end{align}

\fussy

$(\boldsymbol\alpha_u, \boldsymbol\beta_u)$ and $(\boldsymbol\delta_u, \boldsymbol\gamma_u)$ are optimized using a truncated Newton method \cite{NocedalWright2006}.

\section{Discriminative Classification with Fisher Kernels}
\label{Sec:Fisher_kernels}
Fisher kernels~\cite{jaakkola1999} provide a commonly used framework that exploits generative probabilistic models as a representation of instances within discriminative classifiers. Specifically, the Fisher kernel approach involves a feature mapping of structured input---for instance, sequential input---by a projection into the gradient space of a generative probabilistic model that is previously fit on the training data via maximum likelihood. 
We use the generative probabilistic model developed in Section~\ref{Sec:Generative_Model_Lexical_Features} to map scanpaths and lexical features into feature vectors $\mathbf{g}$. 
The Fisher score representation $\mathbf{g}$ for a scanpath $\mathbf{S}$ is the gradient of the log likelihood of $\mathbf{S}$ with respect to the model parameters, evaluated at the maximum likelihood estimate.

\subsection{Fisher Kernel Function}
The Fisher kernel function $K$ calculates the similarity of two scanpaths $\mathbf{S}_i$, $\mathbf{S}_j$ as the inner product of their Fisher score representations $\mathbf{g}_i$ and $\mathbf{g}_j$, relative to the Riemannian metric that is given by the inverse of the Fisher information matrix $\mathbf{I}$.
\begin{definition}[Fisher kernel function of model with lexical features]
Let $\btheta^*$ be the maximum likelihood estimate of the model defined in Section~\ref{Sec:Generative_Model_Lexical_Features} on all training data.
Let $\mathbf{S}_i$, $\mathbf{S}_j$ denote scanpaths on texts $\bX_i$, $\bX_j$. The fisher kernel between $\bS_i$, $\bS_j$ is
\begin{equation*}
K((\bS_i,\bX_i),(\bS_j,\bX_j) = \bg_i^\top \mathbf{I}^{-1} \bg_j
\end{equation*}
where $\bg_i = \left. \nabla_{\btheta} p(\bS_i|\bX_i,\btheta) \right|_{\theta=\theta^*}$ and we employ the empirical version of the Fisher information 
matrix given by $\mathbf{I} = \frac{1}{N} \sum_{i = 1}^N \mathbf{g}_i \mathbf{g}_i^\top.$
The gradient of the log-likelihood function is derived in Proposition~\ref{prop:gradient}.
\end{definition}

\sloppy
\begin{proposition}[Gradient of log-likelihood of generative model with lexical features] \label{prop:gradient}
Let $\bS = ((q_1, d_1),\ldots,(q_T,d_T))$ denote a scanpath obtained on text $\bX$. 
Let $a_1,...,a_T$ denote the saccade amplitudes, and $u_1,...,u_T$ denote the saccade types in $\bS$. 
Define for $u \in \{1,2,3,4,5\}$ the set $\{i^{(u)}_1,...,i^{(u)}_{K_u}\} = \{i \in \{1,...,T\}| u_i = u \}$.
Let \mbox{$\ba_u = (|a_{i^{(u)}_1}|,...,|a_{i^{(u)}_{K_u}}|)^\top$}, \mbox{$\bd_u = (d_{i^{(u)}_1},...,d_{i^{(u)}_{K_u}})^\top$},
and $\bW_u$ the $K_u \times M$ matrix with row vectors $\bw^\top_{i^{(u)}_k}$ for $1 \leq k \leq K_u$.
Then the gradient of the logarithmic likelihood of the model defined in Section~\ref{Sec:Generative_Model_Lexical_Features} is 
\begin{equation*}
\bg=\nabla_{\btheta}  \ln p(\bS|\bX,\btheta) = (\bar\bg_1^\top,\bar\bg_2^\top,\bar\bg_3^\top,\bar\bg_4^\top,\bar\bg_5^\top)^\top
\end{equation*}
where for $u \in \{1,2,3,4,5\}$
\begin{equation*}
\bar\bg_u =\begin{pmatrix}
\pi_u^{-1} K_u\\
\bW_u^\top \Big( \exp(\bW_u \balpha_u) \odot \Big( \ln(\ba_u) - \psi(\exp(\bW_u \balpha_u))-\bW_u \bbeta_u\Big)\Big)\\
\bW_u^\top \Big( \ba_u  \exp(-\bW_u \bbeta_u) -  \exp(\bW_u \balpha_u)\Big)\\
\bW_u^\top \Big( \exp(\bW_u \bdelta_u) \odot \Big( \ln(\bd_u) - \psi(\exp(\bW_u \bdelta_u))-\bW_u \bgamma_u\Big)\Big)\\
\bW_u^\top \Big( \bd_u \exp(-\bW_u \bgamma_u) -  \exp(\bW_u \bdelta_u)\Big)
\end{pmatrix}
\end{equation*}
and $\odot$ denotes the Hadamard product.
\end{proposition}
A proof of Proposition~\ref{prop:gradient} is given in the appendix.
\fussy

\subsection{Applying the Fisher Kernel to Identification and Text Comprehension}
Applying the Fisher Kernel to both prediction problems first requires to estimate the parameters of the generative model parameters on the training data. Note that we fit a global model, instead of class-specific models.
In both prediction problems, we treat the scanpaths of each single line of text as an instance, and 
train a dual SVM with the resulting Fisher kernel.
At application time, the scanpath of a text that is comprised of multiple lines is processed as multiple instances by the Fisher SVM. In order to obtain one decision-function value for the entire text, we average the decision-function values of all individual lines.

\section{Empirical Study}
\label{sec:empirical_study}
\subsection{Data collection}
\subsubsection{Experimental design and materials}
We let a group of 62 advanced and first-semester students read a total of 12 scientific texts on biology (6 texts) and on physics adopted from various German language textbooks \cite{Demtroeder2,Demtroeder3,Demtroeder4,MolBio,Oekologie,Genetik,ZellMolbio}. 
All students are native speakers of German with normal or corrected-to-normal vision and are majoring in either physics or biology.
We determine each reader's comprehension of each text by presenting three comprehension questions after each text. All questions are multiple-choice questions with always one out of four options being correct. Texts have 158 words on average (minimally 126 and maximally 180).

\subsubsection{Technical set-up and Procedure}
 Participants' eye movements are recorded with an SR Research Eyelink 1000 eyetracker (right eye monocular tracking) at a sampling rate of 1000~Hz using a desktop mounted camera system with a 35~mm lens and head stabilization. 
 After setting up the camera and familiarizing the participant with the procedure, the twelve texts are presented in randomized order. Each text fits onto a single screen. We impose no restrictions  regarding the time spent on reading one text. 
 After each text, three comprehension  questions are presented on separate screens together with  4 multiple choice options. 
 Participants cannot backtrack to the text or previous questions, or undo an answer.
The total duration of the experiment is approximately 90 minutes; participants were paid for participating. 

\subsubsection{Lexical features}
\label{lexicalFeats}
Lexical frequency and word length are well known to affect a reader's fixation durations and saccadic behavior, such as whether a word is skipped or a regressive saccade is initiated \cite{RaynerMcConkie1976,RaynerDuffy1986,InhoffRayner1986,Kliegl2003,Kliegl2004,Juhasz2008}. Hence, for each word of the stimuli, we extract different kinds of word frequency and word length measures using dlexDB \cite{dlex,Heister2011}, which is based on the reference corpus underlying the  Digital Dictionary of the German Language (DWDS) corpus \cite{dwds}. Specifically, we extract type frequency (\ie the number of occurrences of a type in the corpus per million tokens), annotated type frequency (\ie the number of occurrences of a unique combination of a type, its part-of-speech, and its lemma in the corpus per million tokens), lemma frequency (\ie the total number of occurrences of types associated with this lemma in the corpus per million tokens), document frequency (\ie  the number of documents with at least one occurrence of this type per 10,000 documents), type length in number of characters, type length in number of syllables, and lemma length in number of characters.  All corpus-based features are log-transformed and z-score normalized. Moreover, we tag each word with the following binary lexical features: whether the word is a technical term,  a technical term from physics, a technical term from biology, an abbreviation, the first word of a sentence.

\subsection{Reference Methods}

We compare the {\em Fisher SVM with lexical features} to several reference methods. The first natural baseline is the {\em generative model with lexical features} developed in Section~\ref{Sec:Generative_Model_Lexical_Features}; this comparison allows us to measure the merit of the discriminative Fisher kernel compared to the underlying generative model. The next baseline is the {\em Fisher SVM without lexical features}---that is, an SVM with the Fisher kernel derived from the generative model described in Section~\ref{sec:landwehr_model}. We compare this discriminative model to the full {\em generative model (Landwehr et al., 2014)}~\cite{Landwehr2014} without lexical features and without the simplification introduced in Section~\ref{sec:landwehr_model}. 

The current gold-standard model for reader identification is the model of {\em Abdelwahab et al., 2016}~\cite{AbdelwahabEtAl2016}. Note that no Fisher kernel can be derived from this non-parametric generative model for lack of explicit model parameters. Since this model has been shown to outperform all previous approaches~\cite{Holland2011,Landwehr2014}, we exclude~\cite{Holland2011} from our comparison.

\subsection{Experimental Setting}

For reader identification, data are split along texts, so that the same text does not appear in training and test data. We conduct a leave-one-text-out cross-validation protocol: The models are trained on 11 texts per reader and a reader is identified on the left-out text. Identification accuracy is averaged across the resulting 12 training- and test-splits and is studied as a function of the number of text lines read at test time. 

For text comprehension, data are split (50/50) across readers and texts, so that neither the same reader nor the same text appears in both training and test data. This setup leads to four train-test splits, across which we average the classification accuracy. 

For both problem settings, we execute another nested cross-validation inside the top-level cross-validation in which we tune the hyperparameters of all learning methods (\eg regularization parameters of the SVM and the linear model for lexical features and parameter $\alpha$ of the non-parametric method of Abdelwahab et al.) by grid search. We also perform feature subset selection on vector $\bw_t$ by backward elimination in this inner cross-validation step. The nested cross-validation protocol ensures that all hyperparameters are tuned on the training part of the data.

\begin{figure}[h]
	\centering
	\includegraphics[width=0.6\textwidth]{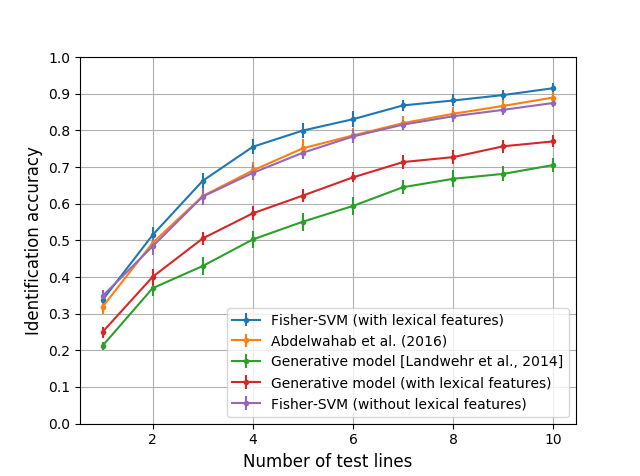}
	\caption{Identification accuracy as a function of lines read at test time; error bars show the standard error. 
	}
	\label{fig:identification accuracy}
\end{figure}

\subsection{Reader Identification}

\begin{figure}[h]
	\centering
	\includegraphics[width=0.6\textwidth]{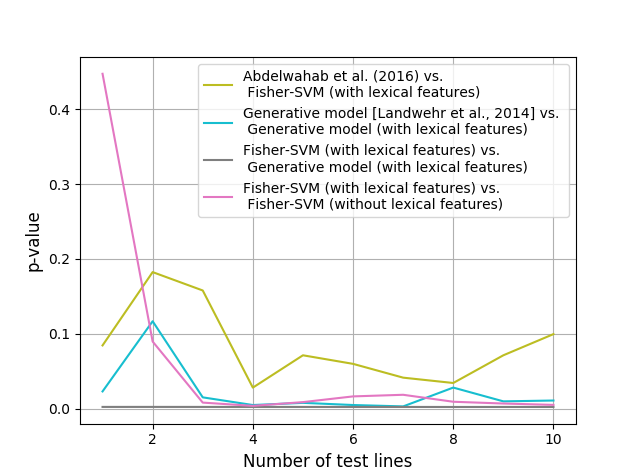}
	\caption{$p$-values of a Wilcoxon signed-rank test for comparison of  model pairs}.
	\label{fig:p-value identification accuracy}
\end{figure}

We measure the percentage of correctly identified readers from the set of 62 readers.  
Figure \ref{fig:identification accuracy} shows the identification accuracy for the different models. 
The Fisher-SVM achieves an identification accuracy of up to $91.53\%$ and outperforms the other evaluated models. 
Figure~\ref{fig:p-value identification accuracy} shows the $p$-value of a Wilcoxon signed-rank test for a comparison of several pairs of methods. 
We conclude that the Fisher-SVM with lexical features  outperforms Abdelwahab significantly ($p<0.05$) for 4 and 8 lines read, the Fisher-SVM with lexical features always outperforms the Fisher-SVM without lexical features, the Fisher-SVM always outperforms the underlying generative model, and the generative model with lexical features outperforms the generative model of Landwehr at al.~without lexical features for 3 or more lines read.
Including lexical features significantly improves the generative model by Landwehr et al.~\cite{Landwehr2014}, as well as the Fisher-SVM.

\subsubsection{Execution Time}

\begin{figure}[h]
	\centering
	\includegraphics[width=0.53\textwidth]{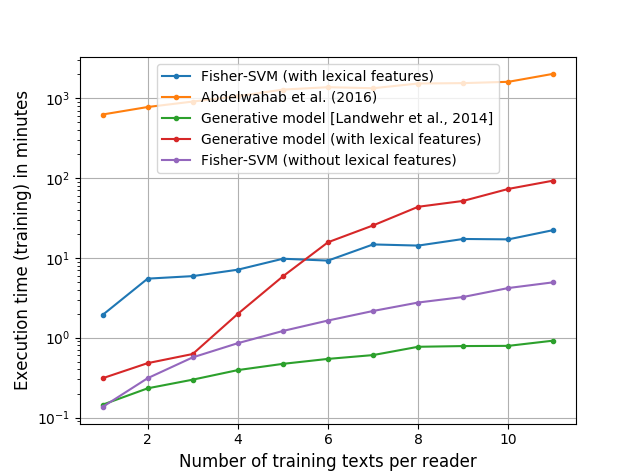}
	\caption{Elapsed execution time of training the models for reader identification, as a function of the number of used training texts per reader on a single ten-core CPU (Intel Xeon E5-2640, 2.40GHz).}\label{fig:times}
\end{figure}

We compare the time required to train reader-identification models for all methods under investigation as a function of the number of training texts per reader. Figure~\ref{fig:times} shows that training the nonparametric model of Abdelwahab et al.~is one to three orders of magnitude slower than all other models. The model of Landwehr et al.~uses a quasi-Newton method to fit the gamma distributions, the generative model with lexical features additionally fits several linear models. Generative models are fit for each reader. By contrast, the Fisher kernel requires fitting one single model to all data and training a linear model; this turns out to be faster in some cases.

\subsubsection{Text Comprehension}
After reading a text, each subject answers three text comprehension questions. 
We study a binary classification problem where one class corresponds to zero or one correct answers and the other class to two or three correct answers. 


Table \ref{tab:results text-comprehension} shows the classification accuracies of the evaluated models\footnote{The main memory requirement of the model of Abdelwahab et al.~is quadratic in the number of instances per class; we had to discard 80\% of the data at random for this problem}. No methods exceeds the classification accuracy of a model that always predicts the majority class. The discriminative models minimize the hinge loss---which is an upper bound of the zero-one loss---and reach the minimal loss by almost always predicting the majority class. The generative models are not trained to minimize any classification loss at all. They fall far short of the accuracy of the majority class but attain an AUC that is marginally above random guessing. The AUC of all three models is significantly higher than 0.5 ($p<0.05$, paired $t$-test). In order to validate this interpretation, we additionally train the Fisher SVM on a class-balanced data subset; with balanced classes, the Fisher SVM cannot minimize the loss without also increasing the AUC. Here, the Fisher SVM achieves an AUC of $0.54\pm 0.03$ which is significantly higher than 0.5.
We conclude that estimating the level of text comprehension is a difficult problem that cannot be solved at any useful level by any of the models under investigation.

\begin{table}[H]
	\caption{Classification accuracy and AUC $\pm$ standard error for text comprehension.} 
	\label{tab:results text-comprehension}
	\centering
	\begin{tabular}{l|l|l}
		Method											&Classification accuracy	& AUC\\ \hline \hline
		Fisher-SVM (lexical features)					& $0.6866 \pm 0.0615$	& $0.5071 \pm 0.0282$\\ \hline
		Fisher-SVM (without lexical features)			& $0.6529 \pm 0.0708$	& $0.5181 \pm 0.0339$	\\ \hline
		Abdelwahab et al. (2016)						& $0.5954 \pm 0.0232$ & $0.5403 \pm 0.0272$	\\ \hline
		Generative model (with lexical features)		& $0.5273 \pm 0.0261$	& $0.5500 \pm 0.0293$\\ \hline
		Generative model [Landwehr et al., 2014]		& $0.5206 \pm 0.0207$	& $0.5555 \pm 0.0120$\\ \hline
		Majority class									& $0.7014 \pm 0.0547$	& 0.5\\ \hline
	\end{tabular}
\end{table}

\section{Conclusions}
We developed a discriminative model for the classification of scanpaths in reading. The aim was to i) predict the readers' identity, and ii) their level of text comprehension. To this end, we built on the work of~\cite{Landwehr2014} and developed a generative graphical model of scanpaths that  takes into account  lexical features of the fixated word, derived a Fisher representation of scanpaths from this model, and subsequently used this Fisher kernel to classify the data using an SVM.
We collected eye-tracking data of 62 readers who read 12 scientific texts and answered comprehension questions for each text. 

We can conclude that the inclusion of lexical features leads to a significant improvement compared to the original generative model~\cite{Landwehr2014}, and that a discriminative model using a Fisher kernel gives an additional considerable improvement over the generative model. We conclude that this model significantly outperforms the semiparametric model of \cite{AbdelwahabEtAl2016} in some cases, which, to the best of our knowledge, is the best published biometric model that is based on eye movements. None of the considered models was able to reliably predict reading comprehension from a reader's eye movements. 

\appendix
\section{Appendix}
\begin{proof}[Proposition~\ref{prop:gradient}]
As discussed in Section~\ref{sec:parameter_estimation}, the likelihood factorizes as
\begin{equation*}
\ln p(\bS|\bX,\btheta) = \sum_{t=1}^{T} \ln \mathrm{Mult}(u_t| \bpi) + \sum_{t=1}^{T} \ln p(a_t | u_t,\bw_t,\balpha,\bbeta) 
+ \sum_{t=1}^{T} \ln p(d_t | u_t,\bw_t,\bgamma,\bdelta). 
\end{equation*}
For the multinomial distribution, 
\begin{equation*}
\sum\nolimits_{t=1}^{T} \ln \mathrm{Mult}(u_t| \bpi) = \ln \frac{T!}{\prod\nolimits_{u=1}^{5} K_u!} + \sum\nolimits_{u=1}^5 K_u \ln \pi_u
\end{equation*}
and thus for $u \in \{1,2,3,4,5\}$, we have that 
$\frac{\partial \ln p(\bS|\bX,\btheta)}{ \partial \pi_u} = \frac{K_u}{\pi_u}.$
Since the likelihoods of the saccade amplitudes and the fixation durations are analogous (see Equations~\ref{eq:amplitude_positive_feat}--\ref{eq:duration_feat}), we only derive the gradient of the amplitude likelihood. 
As discussed in Section~\ref{sec:parameter_estimation} (Equation~\ref{eq:factorization_over_u}), the likelihood of saccade amplitudes and fixation durations further 
factorizes over the different saccade types $u$.
Therefore, if $\alpha_{u,m}$ denotes the $m$-th entry of parameter vector $\balpha_u$, its partial derivative is
\begin{align}
\MoveEqLeft \frac{\partial} {\partial \alpha_{u,m}} \sum_{1 \leq t \leq T:u_t=u} \ln p(a_t | u_t,\bw_t,\balpha,\bbeta) \notag\\
&= \frac{\partial} {\partial \alpha_{u,m}}  \sum_{1 \leq t \leq T:u_t=u} \ln \mathcal{G}(|a_t|| \exp(\balpha_u^\top \mathbf{w}^{(i)}_t), \exp(\bbeta_u^\top \mathbf{w}^{(i)}_t))\notag\\
& = \frac{\partial }{\partial \alpha_{u,m}}\sum_{1 \leq t \leq T:u_t=u} 
\ln \left(|a_t|^{{\operatorname{exp}(\boldsymbol\alpha_u^\top {\bf w}_t)}-1}\operatorname{exp}({-\frac{|a_t|}{{\operatorname{exp}(\boldsymbol\beta_u^\top  {\bf w}_t)}})}\right) \notag\\
&\hspace{30mm} -\ln(\Gamma({\operatorname{exp}(\boldsymbol\alpha_u^\top  {\bf w}_t)}){\operatorname{exp}(\boldsymbol\beta_u^\top {\bf w}_t)}^{\operatorname{exp}(\boldsymbol\alpha_u^\top {\bf w}_t)})\notag\\
&=\sum\limits_{\substack{1 \leq t \leq T:u_t=u}} w_{t,m} \operatorname{exp}(\boldsymbol\alpha_u^\top  {\bf w}_t) \Big( \ln(|a_t|) - \psi(\operatorname{exp}(\boldsymbol\alpha_u^\top  {\bf w}_t))-\boldsymbol\beta_u^\top  {\bf w}_t\Big)\label{eq:gradalpha}.
\end{align}

Moreover, if $\beta_{u,m}$ denotes the $m$-th entry of parameter vector $\bbeta_u$, 
\begin{align}
\MoveEqLeft\frac{\partial} {\partial \beta_{u,m}} \sum_{1 \leq t \leq T:u_t=u} \ln p(|a_t| | u_t,\bw_t,\balpha,\bbeta) \notag\\
& = \frac{\partial} {\partial \beta_{u,m}}  \sum_{1 \leq t \leq T:u_t=u} \ln \mathcal{G}(|a_t|| \exp(\balpha_u^\top \mathbf{w}^{(i)}_t), \exp(\bbeta_u^\top \mathbf{w}^{(i)}_t))\notag\\
& = \frac{\partial }{\partial \beta_{u,m}}\sum_{1 \leq t \leq T:u_t=u} 
\ln \left(|a_t|^{{\operatorname{exp}(\boldsymbol\alpha_u^\top {\bf w}_t)}-1}\operatorname{exp}({-\frac{|a_t|}{{\operatorname{exp}(\boldsymbol\beta_u^\top  {\bf w}_t)}})}\right) \notag\\
&\hspace{30mm} -\ln(\Gamma({\operatorname{exp}(\boldsymbol\alpha_u^\top  {\bf w}_t)}){\operatorname{exp}(\boldsymbol\beta_u^\top {\bf w}_t)}^{\operatorname{exp}(\boldsymbol\alpha_u^\top {\bf w}_t)})\notag\\
&=\sum\limits_{\substack{1 \leq t \leq T:u_t=u}} w_{t,m}\Big(|a_t| \operatorname{exp}(-\boldsymbol\beta_u^\top {\bf w_t}) -  \operatorname{exp}(\boldsymbol\alpha_u^\top {\bf w_t})\Big).\label{eq:gradbeta}
\end{align}
In Equations~\ref{eq:gradalpha} and~\ref{eq:gradbeta} we have exploited that the derivative of the log-gamma function is given by the digamma function $\psi$---\ie $\frac{d}{dx}\ln \Gamma(x)=\psi(x)$. The claim now follows from straightforward calculation.
\end{proof}

\subsection*{Acknowledgment}
This work was partially funded by the German Science Foundation under grants SFB1294, SFB1287, and LA3270/1-1, and by the German Federal Ministry of Research and Education under grant 16DII116-DII.

\bibliographystyle{splncs}
\bibliography{ecml2018}

\begin{thebibliography}{10}

\bibitem{kliegl2006}
Kliegl, R., Nuthmann, A., Engbert, R.:
\newblock Tracking the mind during reading: The influence of past, present, and
  future words on fixation durations.
\newblock Journal of Experimental Psychology: General, 135(1) (2006)  12–35

\bibitem{Rayner1998}
Rayner, K.:
\newblock Eye movements in reading and information processing: 20 years of
  research.
\newblock Psychological Bulletin \textbf{124}(3) (1998)  372--422

\bibitem{erdmann1898}
Erdmann, B., Dodge, R.:
\newblock Psychologische Untersuchungen {\"u}ber das Lesen auf experimenteller
  Grundlage.
\newblock Books on Demand (1898)

\bibitem{huey1908}
Huey, E.B.:
\newblock The psychology and pedagogy of reading.
\newblock The Macmillan Company (1908)

\bibitem{Afflerbach2015}
Afflerbach, P., ed.:
\newblock Handbook of Individual Differences in Reading.
\newblock Routledge (2015)

\bibitem{Holland2011}
Holland, C., Komogortsev, O.V.:
\newblock Biometric identification via eye movement scanpaths in reading.
\newblock In: Proceedings of the 2011 International Joint Conference on
  Biometrics. IJCB '11, Washington, DC, IEEE (2011)  1--8

\bibitem{Landwehr2014}
Landwehr, N., Arzt, S., Scheffer, T., Kliegl, R.:
\newblock A model of individual differences in gaze control during reading.
\newblock In: EMNLP. (2014)  1810--1815

\bibitem{AbdelwahabEtAl2016}
Abdelwahab, A., Kliegl, R., Landwehr, N.:
\newblock A semiparametric model for bayesian reader identification.
\newblock In: Proceedings of the 2016 Conference on Empirical Methods in
  Natural Language Processing (EMNLP-2016), Austin, TX (2016)

\bibitem{JustCarpenter1980}
Just, M.A., Carpenter, P.A.:
\newblock A theory of reading: From eye fixations to comprehension.
\newblock Psychological Review \textbf{87}(4) (1980)  329--354

\bibitem{Frazier1982}
Frazier, L., Rayner, K.:
\newblock Making and correcting errors during sentence comprehension: {E}ye
  movements in the analysis of structurally ambiguous sentences.
\newblock Cognitive Psychology \textbf{14}(2) (1982)  178--210

\bibitem{Staub2007}
Staub, A., Rayner, K., Pollatsek, A., Hy\"on\"a, J., Majewski, H.:
\newblock The time course of plausibility effects on eye movements in reading:
  {E}vidence from noun-noun compounds.
\newblock Journal of Experimental Psychology: {L}earning, Memory, and Cognition
  \textbf{33}(6) (2007)  1162--1169

\bibitem{Kaakinen2007}
Kaakinen, J.K., Hy\"on\"a, J.:
\newblock Perspective effects in repeated reading: {A}n eye movement study.
\newblock Memory and Cognition \textbf{35} (2007)  1323--1336

\bibitem{Rayner2006}
Rayner, K., Chace, K.H., Slattery, T.J., Ashby, J.:
\newblock Eye movements as reflections of comprehension processes in reading.
\newblock Scientific Studies of Reading \textbf{10}(3) (2006)  241--255

\bibitem{Rayner1994}
Rayner, K., Sereno, S.C.:
\newblock Eye movements in reading: {P}sycholinguistic studies.
\newblock In Gernsbacher, M.A., ed.: Handbook of Psycholinguistics.
\newblock Academic Press, San Diego (1994)  57--81

\bibitem{Clifton2007}
Clifton, C., Staub, A., Rayner, K.:
\newblock Eye movements in reading words and sentences.
\newblock In {Van Gompel}, R.P., Fischer, M.H., Murray, W.S., Hill, R.L., eds.:
  Eye Movements: {A} Window on Mind and Brain.
\newblock Elsevier, Oxford, UK (2007)  341--372

\bibitem{Jacobson1979}
Jacobson, J.Z., Dodwell, P.C.:
\newblock Saccadic eye movements during reading.
\newblock Brain and Language \textbf{8} (1979)  303--314

\bibitem{RaynerPollatsek1989}
Rayner, K., Pollatsek, A.:
\newblock The Psychology of Reading.
\newblock Prentice Hall, Englewood Cliffs, NJ (1989)

\bibitem{Rayner2004}
Rayner, K., Warren, T., Juhasz, B.J., Liversedge, S.P.:
\newblock The effect of plausibility on eye movements in reading.
\newblock Journal of Experimental Psychology: {L}earning, Memory, and Cognition
  \textbf{30} (2004)  1290--1301

\bibitem{Warren2008}
Warren, T., McConnell, K., Rayner, K.:
\newblock Effects of context on eye movements when reading about possible and
  impossible events.
\newblock Journal of Experimental Psychology: {L}earning, Memory, and Cognition
  \textbf{34} (2008)  1001--1007

\bibitem{Copeland2013}
Copeland, L., Gedeon, T.:
\newblock Measuring reading comprehension using eye movements.
\newblock In: 4th IEEE International Conference on Cognitive (CogInfoCom 2013).
  (2013)  791--796

\bibitem{Copeland2014b}
Copeland, L., Gedeon, T., Mendis, S.:
\newblock Predicting reading comprehension scores from eye movements using
  artificial neural networks and fuzzy output error.
\newblock Artificial Intelligence Research \textbf{3}(3) (2014)  35--48

\bibitem{Copeland2014a}
Copeland, L., Gedeon, T., Mendis, S.:
\newblock Fuzzy output error as the performance function for training
  artificial neural networks to predict reading comprehension from eye gaze.
\newblock In Loo, C., Keem~Siah, Y., Wong, K., Beng~Jin, A., Huang, K., eds.:
  The 21st International Conference on Neural Information Processing 2014
  (ICONIP 2014). Volume~1 of Lecture Notes in Computer Science (LNCS) 8834.
  (2014)  586--593

\bibitem{Underwood1990}
Underwood, G., Hubbard, A., Wilkinson, H.:
\newblock Eye fixations predict reading comprehension: {T}he relationships
  between reading skill, reading speed, and visual inspection.
\newblock Language and Speech \textbf{33}(1) (1990)  69--81

\bibitem{dixon1951studies}
Dixon, W.R.:
\newblock Studies in the psychology of reading.
\newblock In Morse, W.S., Ballantine, P.A., Dixon, W.R., eds.: Univ. of
  Michigan Monographs in Education No. 4.
\newblock Univ. of Michigan Press (1951)

\bibitem{rayner2012psychology}
Rayner, K., Pollatsek, A., Ashby, J., Clifton~Jr, C.:
\newblock Psychology of reading.
\newblock Psychology Press (2012)

\bibitem{komogortsev2010biometric}
Komogortsev, O.V., Jayarathna, S., Aragon, C.R., Mahmoud, M.:
\newblock Biometric identification via an oculomotor plant mathematical model.
\newblock In: Proceedings of the 2010 Symposium on Eye-Tracking Research \&
  Applications. (2010)

\bibitem{rigas2012human}
Rigas, I., Economou, G., Fotopoulos, S.:
\newblock Human eye movements as a trait for biometrical identification.
\newblock In: Proceedings of the IEEE 5th International Conference on
  Biometrics: Theory, Applications and Systems. (2012)

\bibitem{zhang2012on}
Zhang, Y., Juhola, M.:
\newblock On biometric verification of a user by means of eye movement data
  mining.
\newblock In: Proceedings of the 2nd International Conference on Advances in
  Information Mining and Management. (2012)

\bibitem{bednarik2005eye}
Bednarik, R., Kinnunen, T., Mihaila, A., Fr{\"a}nti, P.:
\newblock Eye-movements as a biometric.
\newblock In: Proceedings of the 14th Scandinavian Conference on Image
  Analysis. (2005)

\bibitem{rigas2012biometric}
Rigas, I., Economou, G., Fotopoulos, S.:
\newblock Biometric identification based on the eye movements and graph
  matching techniques.
\newblock Pattern Recognition Letters \textbf{33}(6) (2012)

\bibitem{rigas2016biometric}
Rigas, I., Komogortsev, O., Shadmehr, R.:
\newblock Biometric recognition via eye movements: Saccadic vigor and
  acceleration cues.
\newblock ACM Transaction on Applied Perception \textbf{13}(2) (2016)  1--21

\bibitem{MorrisonRayner1981}
Morrison, R.E., Rayner, K.:
\newblock Saccade size in reading depends upon character spaces and not visual
  angle.
\newblock Perception and Psychophysics \textbf{30} (1981)  395--396

\bibitem{NocedalWright2006}
Nocedal, J., Wright, S.J.:
\newblock Numerical Optimization.
\newblock Springer, New York (2006)

\bibitem{jaakkola1999}
Jaakkola, T., Haussler, D.:
\newblock Exploiting generative models in discriminative classifiers.
\newblock In: Advances in neural information processing systems. (1999)
  487--493

\bibitem{Demtroeder2}
Demtr\"oder, W.:
\newblock Experimentalphysik 2: Elektrizität und Optik. 5th edn.
\newblock Springer, Berlin (2009)

\bibitem{Demtroeder3}
Demtr\"oder, W.:
\newblock Experimentalphysik 3: Atome, Moleküle und Festkörper. 4th edn.
\newblock Springer, Berlin (2010)

\bibitem{Demtroeder4}
Demtr\"oder, W.:
\newblock Experimentalphysik 4: Kern-, Teilchen- und Astrophysik. 4th edn.
\newblock Springer, Berlin (2014)

\bibitem{MolBio}
Ableitner, O.:
\newblock Einführung in die Molekularbiologie. Basiswissen für das Arbeiten
  im Labor.
\newblock Springer, Wiesbaden (2014)

\bibitem{Oekologie}
Townsend, C.R., Begon, M., Harper, J.L.:
\newblock \"Okologie.
\newblock Springer, Berlin (2003)

\bibitem{Genetik}
Graw, J.:
\newblock Genetik. 6th edn.
\newblock Springer, Berlin (2015)

\bibitem{ZellMolbio}
Boujard, D., Anselme, B., Cullin, C., Ragu\'en\`es-Nicol, C.:
\newblock Zell- und Molekularbiologie im \"Uberblick.
\newblock Springer, Berlin (2014)

\bibitem{RaynerMcConkie1976}
Rayner, K., McConkie, G.W.:
\newblock What guides a reader's eye movements.
\newblock Vision Research \textbf{16} (1976)  829--837

\bibitem{RaynerDuffy1986}
Rayner, K., Duffy, S.A.:
\newblock Lexical complexity and fixation times in reading: {E}ffects of word
  frequency, verb complexity, and lexical ambiguity.
\newblock Memory and Cognition \textbf{14} (1986)  191--201

\bibitem{InhoffRayner1986}
Rayner, K., Duffy, S.A.:
\newblock Parafoveal word processing during eye fixations in reading: {E}ffects
  of word frequency.
\newblock Perception and Psychophysics \textbf{40} (1986)  431--440

\bibitem{Kliegl2003}
Kliegl, R., Nuthmann, A., Engbert, R.:
\newblock Tracking the mind during reading: {T}he influence of past, present,
  and future words on fixation durations.
\newblock Journal of Experimental Psychology: {G}eneral \textbf{135} (2003)
  12--35

\bibitem{Kliegl2004}
Kliegl, R., Grabner, E., Rolfs, M., Engbert, R.:
\newblock Length, frequency, and predictability effects of words on eye
  movements in reading.
\newblock European Journal of Cognitive Psychology \textbf{16}(1--2) (2004)
  262--284

\bibitem{Juhasz2008}
Juhasz, B.J., White, S.J., Liversedge, S.P., Rayner, K.:
\newblock Eye movements and the use of parafoveal word length information in
  reading.
\newblock Journal of Experimental Psychology: {H}uman Perception and
  Performance \textbf{34} (2008)  1560--1579

\bibitem{dlex}
of~Science, B.B.A., of~Potsdam, U.
\newblock \url{http://dlexdb.de} (2011)

\bibitem{Heister2011}
Heister, J., W\"urzner, K.M., Bubenzer, J., Pohl, E., Hanneforth, T., Geyken,
  A., Kliegl, R.:
\newblock {dlexDB} -- eine lexikalische {D}atenbank f\"ur die psychologische
  und linguistische {F}orschung.
\newblock {P}sychologische {R}undschau \textbf{62}(1) (2011)  10--20

\bibitem{dwds}
Klein, W., Geyken, A., eds.:
\newblock Das Digitale {W}\"orterbuch der deutschen {S}prache ({DWDS}).
\newblock {Berlin-Brandenburg Academy of Science} (2016)
  \url{http://www.dwds.de}.

\end{thebibliography}

\end{document}